\documentclass{article}
\usepackage{spconf,amsmath,graphicx}
\usepackage{amsthm}
\usepackage{amssymb}
\ninept

\usepackage{algorithm,amsfonts}
\usepackage{float}
\usepackage{algpseudocode}
\usepackage{tikz}
\usepackage{pgfplots}
\usepackage{subcaption}
\usepgfplotslibrary{fillbetween}
\usepgfplotslibrary{groupplots}
\input{latex_resources/mysymbol.sty}
\input{latex_resources/juan_defs.sty}
\usepackage{multirow}

\title{Multi-task Bias-Variance Trade-off Through Functional Constraints}
\name{Juan Cervi\~no$^1$, Juan Andr\'es Bazerque$^2$, Miguel Calvo-Fullana$^3$, and Alejandro Ribeiro$^1$ \thanks{Support by NSF CCF 1717120, ARL DCIST CRA under Grant W911NF-17-2-0181 and Theorinet Simons.}}
\address{$^1$Department of Electrical and Systems Engineering, University of Pennsylvania, Philadelphia, USA\\
$^2$Department of Electrical and Computer Engineering, University of Pittsburgh, Pittsburgh, USA\\
$^3$Department of Aeronautics and Astronautics, Massachusetts Institute of Technology, Boston, USA}

\begin{document}
%
\maketitle
\begin{abstract}
Multi-task learning aims to acquire  a set of functions, either regressors or classifiers, that perform well for diverse tasks.
At its core,  the idea behind multi-task learning is to exploit the intrinsic similarity across data sources to aid in the learning process for each individual domain. 
In this paper we draw intuition from the two extreme learning scenarios -- a single function for all tasks, and a task-specific function that ignores the other tasks dependencies -- to  propose a bias-variance trade-off.  
To control the relationship between the variance (given by the number of i.i.d. samples), and the bias (coming from data from other task), we introduce a constrained learning formulation that enforces domain specific solutions to be close to a central function. 
This problem is solved in the dual domain, for which we propose a stochastic primal-dual algorithm. 
Experimental results for a multi-domain classification problem with real data show that the proposed procedure outperforms both the task specific, as well as the single classifiers.
\end{abstract}
\section{Introduction}
\label{sec:intro}

Multi-task learning \cite{caruana1997multitask} and related approaches such as meta-learning \cite{finn2017model,chua2021fine} have recently emerged as popular techniques to handle similarity across data from different domains. The core of multi-task learning resides in exploiting knowledge from closely related sources to improve learning performance across tasks. We are interested in a problem where training data from each domain might be limited. In this scenario, multi-task approaches are able to enhance each individual  performance by sharing relevant information across domains. Operating on finite data encourages formulations that exploit information from related tasks to improve their domain-specific performance. Multi-task learning is particularly useful when the amount of data per source is small,  and its application has shown improvements in practice in the contexts of medical data \cite{ZHOU2021101918}, speech recognition \cite{cai2021speech}, and anomaly detection \cite{georgescu2021anomaly}. 

The challenge in multi-task learning relies on what to share between sources and how to do so \cite{zhang2017survey,vafaeikia2020brief}. Several approaches have been taken in the multi-task learning over the recent years. In the context of deep learning, some works have looked into sharing common representations of the data, which generally translates into the first layers of a neural network \cite{maurer2016benefit,yang2016deep}. Other works, have developed deep architectures with blocks that are shared among tasks \cite{misra2016cross,wallingford2022task} generally by empirically studying which block has a more positive effect on the overall performance. In these works, a part of the architecture is shared, while another one is kept private for each task. 

Measuring the relationship between tasks is also an active area of research in the field of multi-task learning. This line of work focuses on clustering tasks together so as to learn the tasks within a cluster in a combined way \cite{standley2020tasks}. Other methods propose to asses the difference between tasks by measuring disagreements between learned models over the data \cite{ma2018modeling}. Other methods utilize a convex surrogate of the learning objective to model the relationships between tasks \cite{zhang2012convex}. 

In this work, we draw intuition from the two extreme approaches to the multi-task learning problem, (i) disregard the relationship between domains, and doing an agnostic estimation using only the task-specific data, or (ii) treat all data as coming from the same source, and utilizing only one central function for all domains. We recast these two estimators and we define the multi-task bias-variance trade-off between the unbiased task specific estimator (that only utilizes i.i.d. data from a given source), and the centralized biased estimator (with smaller variance given that it utilizes all the samples).
To control the relationship between the variance (given by the number of samples), and the bias (using data from other tasks), we introduce a constrained learning formulation that allows a task specific estimator while keeping the estimator close in the functional domain \cite{kay2008rethinking}. 
Novel in this work, is the use of constrains directly in the function space, in contrast to constraint in the parameter space as in previous work \cite{cervino2019meta,cervino2020multi}. The resulting optimization problem is non-convex. Nonetheless, we exploit recent results in stochastic dual optimization, \cite{chamon_non_convex_losses,chamon2020empirical,chamon2020probably}, and we propose to solve the problem in the dual domain, introducing a cross-learning functional algorithm. Numerical results, studying an image classification problem across different domains show that the proposed functional approach outperforms the previously introduced parametrization-constrained approach, as well as the two extreme cases; the task specific, and the centralized problems.

\section{Multi-task Learning}
\label{sec:MTL}

We consider the problem of optimizing a set of functions that classify data coming from different domains. Instead of finding the classifier for each domain separately as a different task, we attempt to optimize all classifiers jointly. Formally, consider a finite set of tasks $t\in[1,\dots,T]$, and let function $f:\mathcal{X}\times\Theta\to \mathcal{Y}$ be the map between the input space $\mathcal{X}\subset\mathbb{R}^P$, and output space $\mathcal{Y}\subset\mathbb{R}^Q$ parameterized by $\theta\in \Theta\subset\mathbb{R}^S$.  Our objective is to find the parameterizations $\{\theta_t\}$ that minimize the expected loss function $\ell$ over the joint probabilities $p_t(x,y)$,
\begin{align}\label{prob:MTL_agnostic}
   P_{A}^*=\underset{\{\theta_t\}}{\min}  \quad   & \sum_{t=1}^T  \mathbb{E}_{p_t(x,y)}[ \ell\left(y,f(x,\theta_t)\right) ].
\end{align}
In practice, we do not have access to the distributions $p_t(x,y)$ under each task, and therefore the multi-task learning problem as stated in \eqref{prob:MTL_agnostic}, cannot be solved. In this work, we assume that we have a dataset $\mathcal{D}_t$ from each domain $t\in [1,\dots,T]$,  containing labeled samples  $(x_i,y_i)\in(\mathcal{X}$, $\mathcal{Y}),i=1,\dots,N_t$, which are drawn according to the unknown probabilities $p_t(x,y)$. Thus, an empirical version of the multi-task learning problem \eqref{prob:MTL_agnostic} can be written as,

\begin{figure}[t]
\captionsetup[subfigure]{labelformat=empty}
\centering
\begin{subfigure}[b]{.04\columnwidth}
\rotatebox{90}{\quad \small{Art}}
\end{subfigure}
\begin{subfigure}[b]{.15\columnwidth}
\includegraphics[width=1.4cm,height=1.4cm,keepaspectratio]{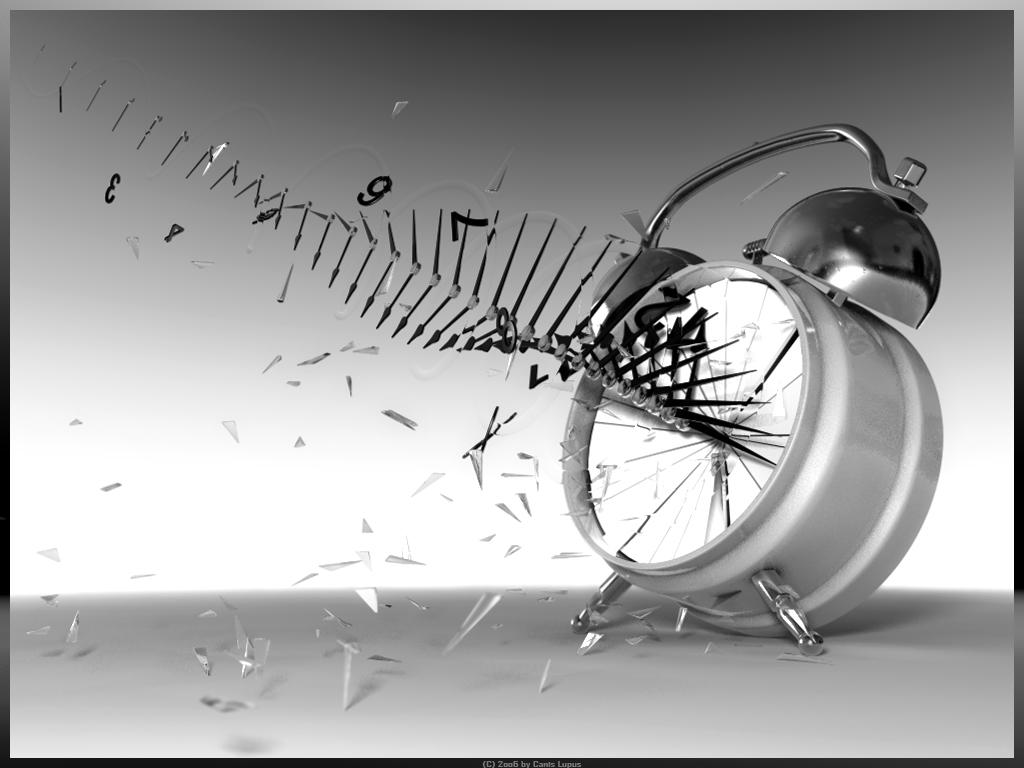}   
\end{subfigure}
\begin{subfigure}[b]{.15\columnwidth}
\includegraphics[width=1.4cm,height=1.4cm,keepaspectratio]{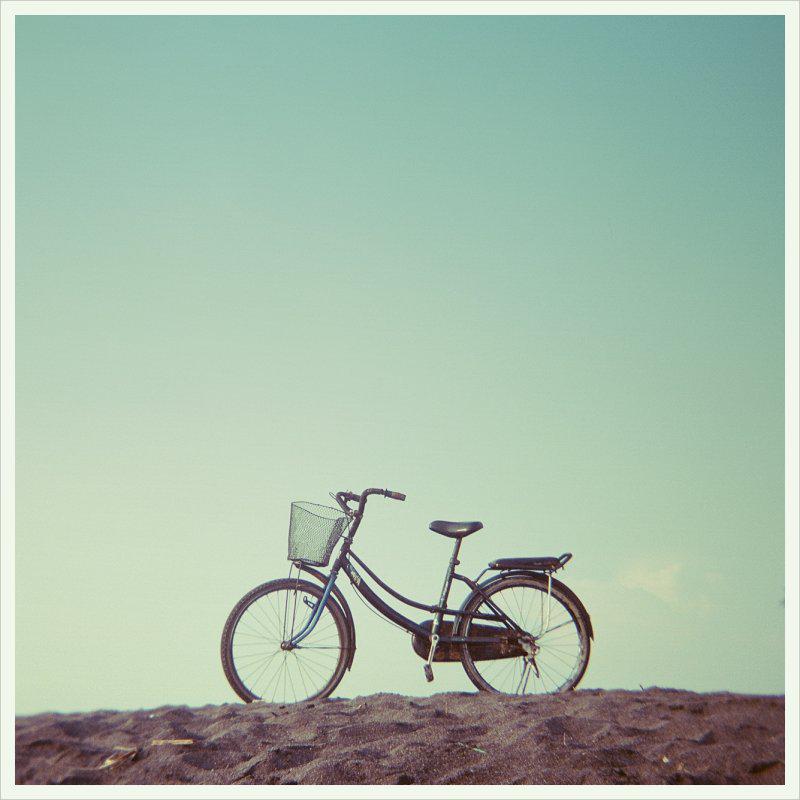}      
\end{subfigure}
\begin{subfigure}[b]{.15\columnwidth}
\includegraphics[width=1.4cm,height=1.4cm,keepaspectratio]{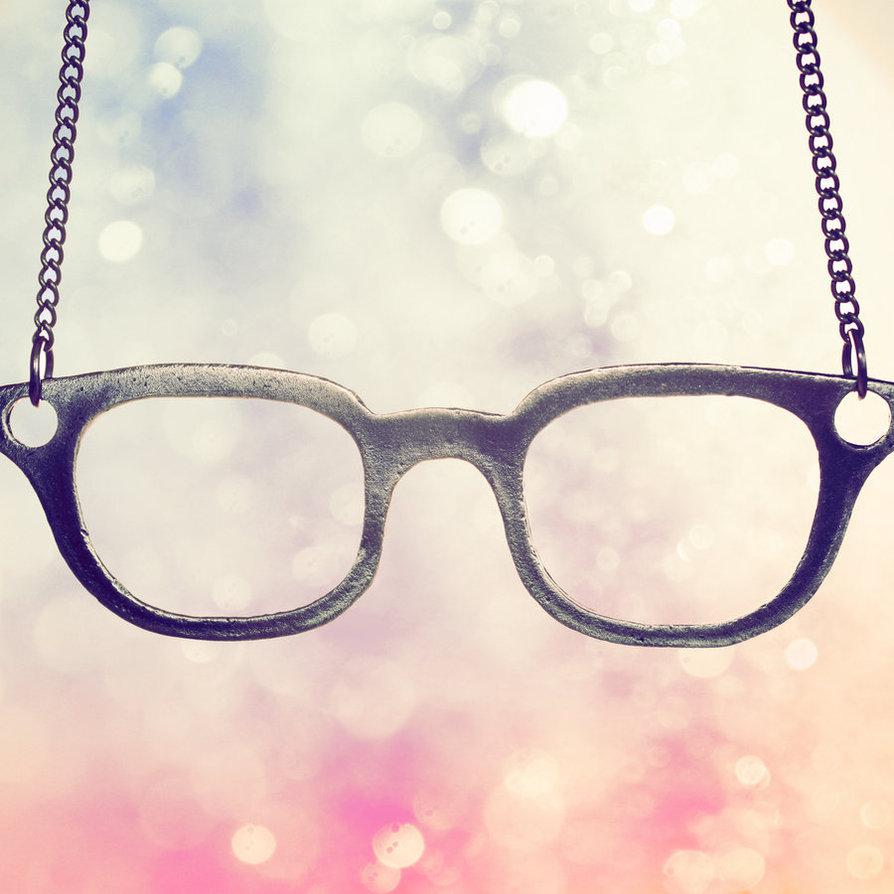}    
\end{subfigure}
\begin{subfigure}[b]{.15\columnwidth}
\includegraphics[width=1.4cm,height=1.4cm,keepaspectratio]{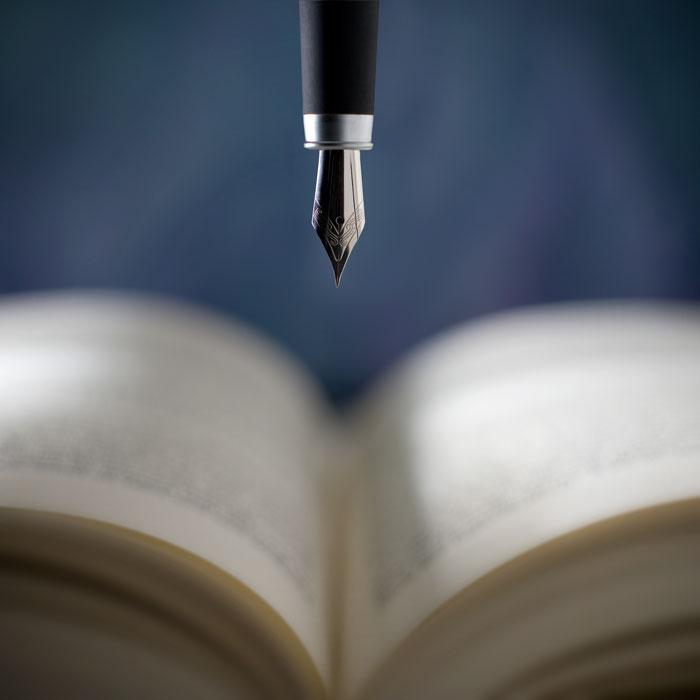}      
\end{subfigure}
\begin{subfigure}[b]{.15\columnwidth}
\includegraphics[width=1.4cm,height=1.4cm,keepaspectratio]{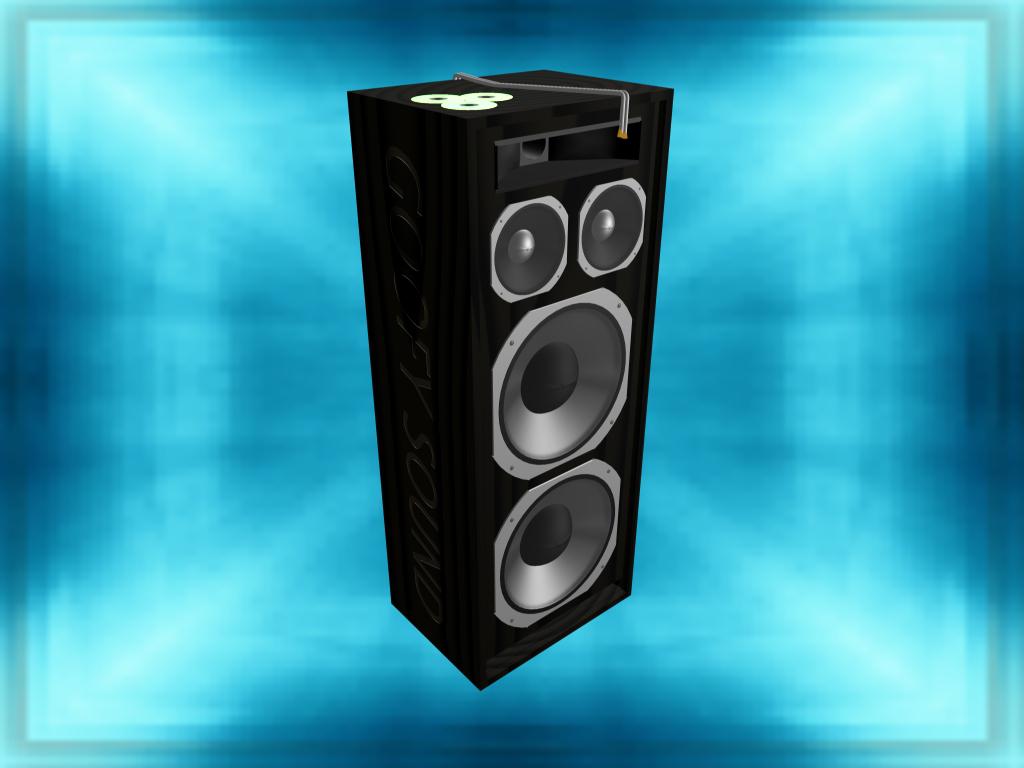}      
\end{subfigure}
\\
\begin{subfigure}[b]{.04\columnwidth}
\rotatebox{90}{\enskip \small{Clipart}}
\end{subfigure}
\begin{subfigure}[b]{.15\columnwidth}
\includegraphics[width=1.4cm,height=1.4cm,keepaspectratio]{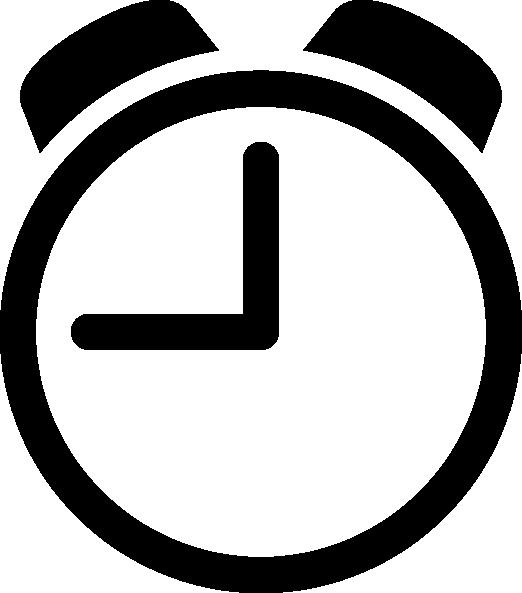}   
\end{subfigure}
\begin{subfigure}[b]{.15\columnwidth}
\includegraphics[width=1.4cm,height=1.4cm,keepaspectratio]{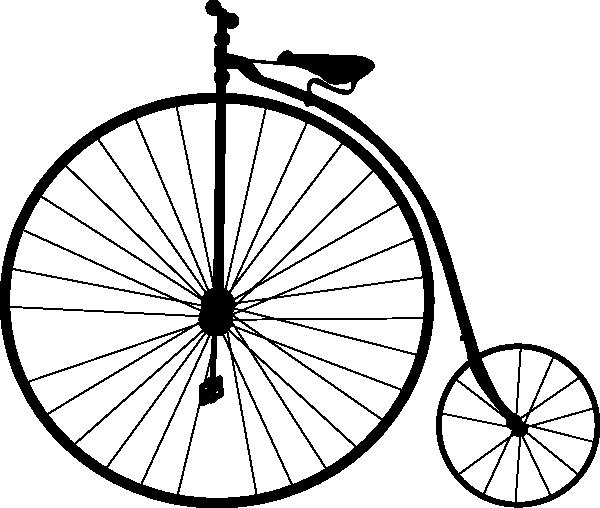}      
\end{subfigure}
\begin{subfigure}[b]{.15\columnwidth}
\includegraphics[width=1.4cm,height=1.4cm,keepaspectratio]{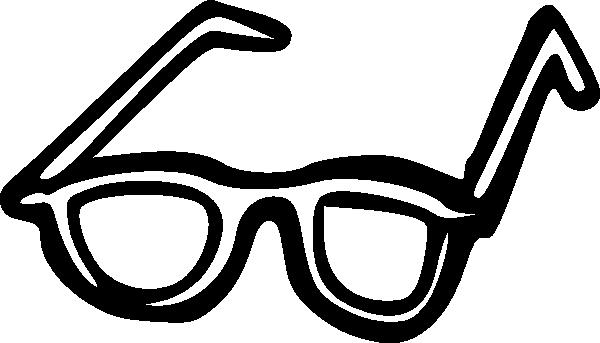}    
\end{subfigure}
\begin{subfigure}[b]{.15\columnwidth}
\includegraphics[width=1.4cm,height=1.4cm,keepaspectratio]{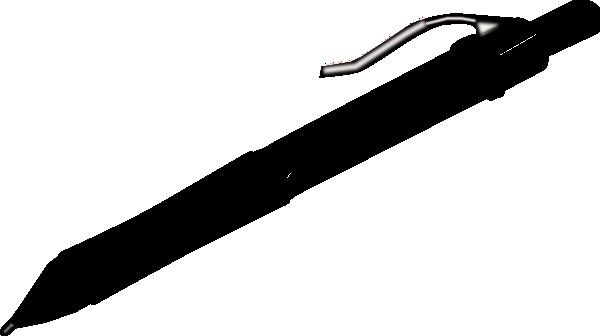}      
\end{subfigure}
\begin{subfigure}[b]{.15\columnwidth}
\includegraphics[width=1.4cm,height=1.4cm,keepaspectratio]{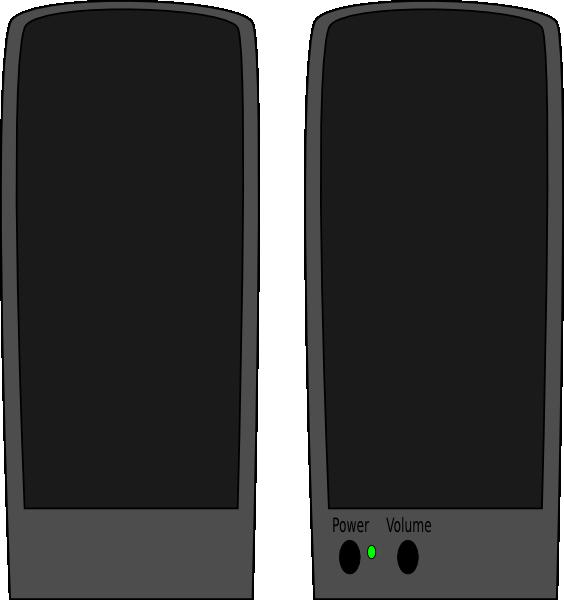}      
\end{subfigure}
\\
\begin{subfigure}[b]{.04\columnwidth}
\rotatebox{90}{\enskip \small{Product}}
\end{subfigure}
\begin{subfigure}[b]{.15\columnwidth}
\includegraphics[width=1.4cm,height=1.4cm,keepaspectratio]{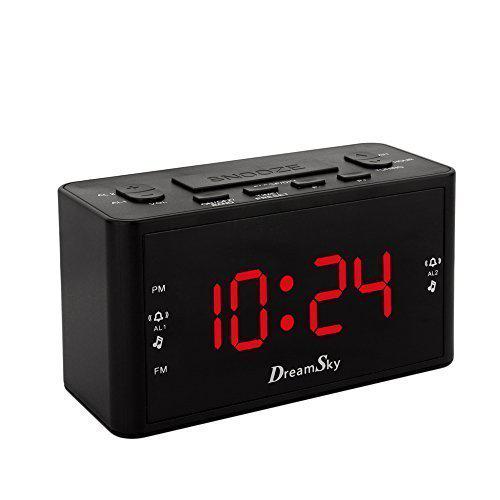}   
\end{subfigure}
\begin{subfigure}[b]{.15\columnwidth}
\includegraphics[width=1.4cm,height=1.4cm,keepaspectratio]{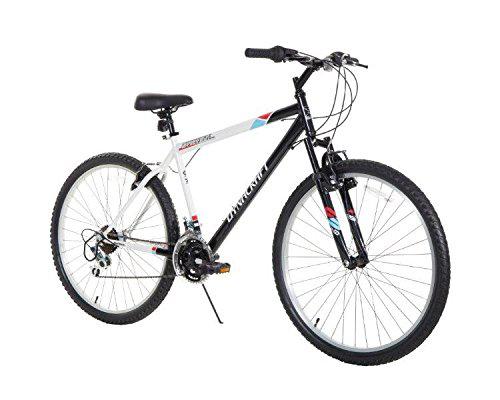}      
\end{subfigure}
\begin{subfigure}[b]{.15\columnwidth}
\includegraphics[width=1.4cm,height=1.4cm,keepaspectratio]{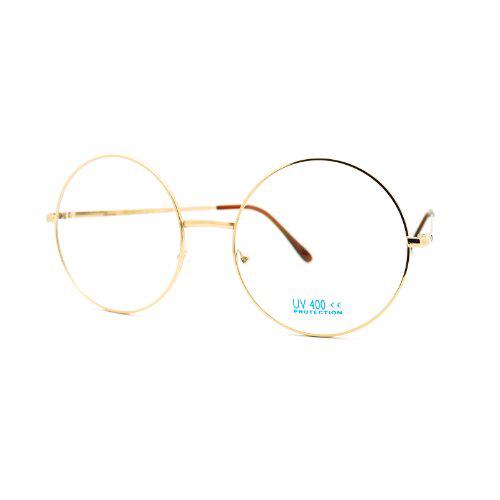}    
\end{subfigure}
\begin{subfigure}[b]{.15\columnwidth}
\includegraphics[width=1.4cm,height=1.4cm,keepaspectratio]{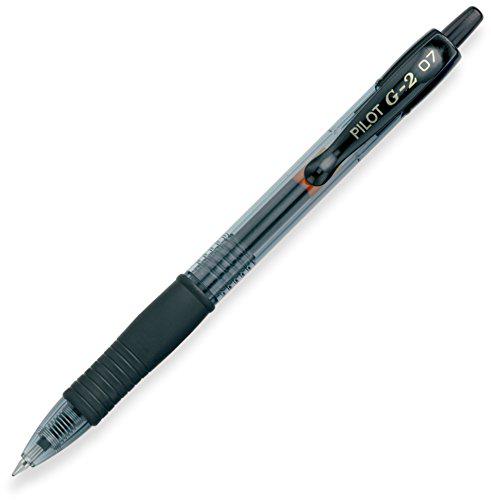}      
\end{subfigure}
\begin{subfigure}[b]{.15\columnwidth}
\includegraphics[width=1.4cm,height=1.4cm,keepaspectratio]{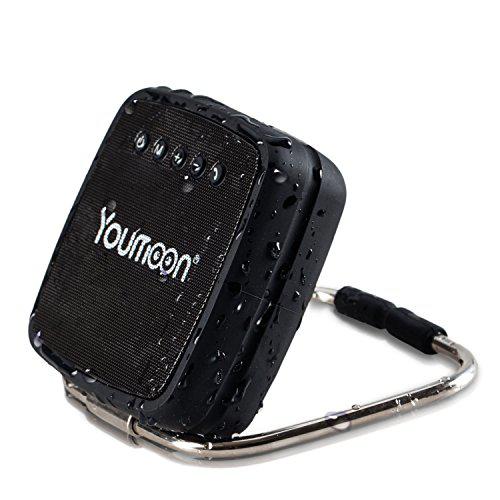}      
\end{subfigure}
\\
\begin{subfigure}[b]{.04\columnwidth}
\rotatebox{90}{\quad\quad \small{World}}
\end{subfigure}
\begin{subfigure}[b]{.15\columnwidth}
\includegraphics[width=1.4cm,height=1.4cm,keepaspectratio]{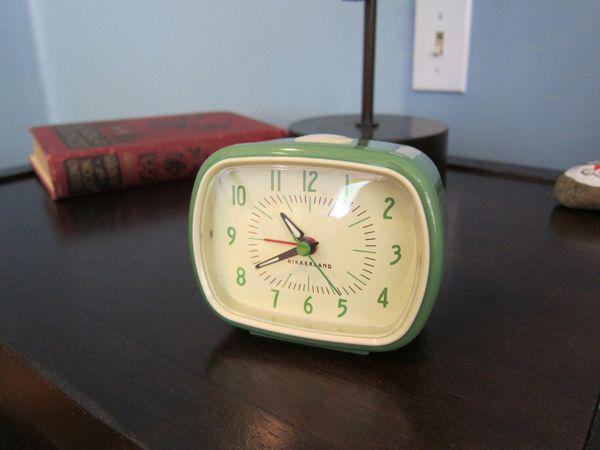}   
\caption{Alarm}
\end{subfigure}
\begin{subfigure}[b]{.15\columnwidth}
\includegraphics[width=1.4cm,height=1.4cm,keepaspectratio]{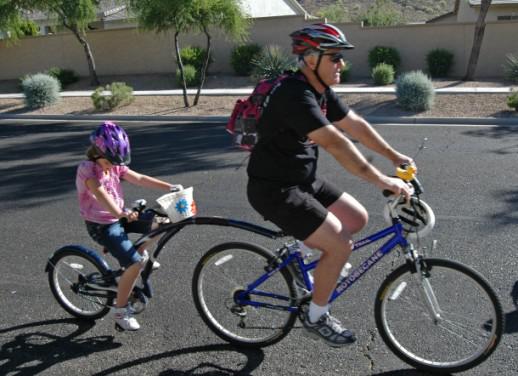}      
\caption{Bike}
\end{subfigure}
\begin{subfigure}[b]{.15\columnwidth}
\includegraphics[width=1.4cm,height=1.4cm,keepaspectratio]{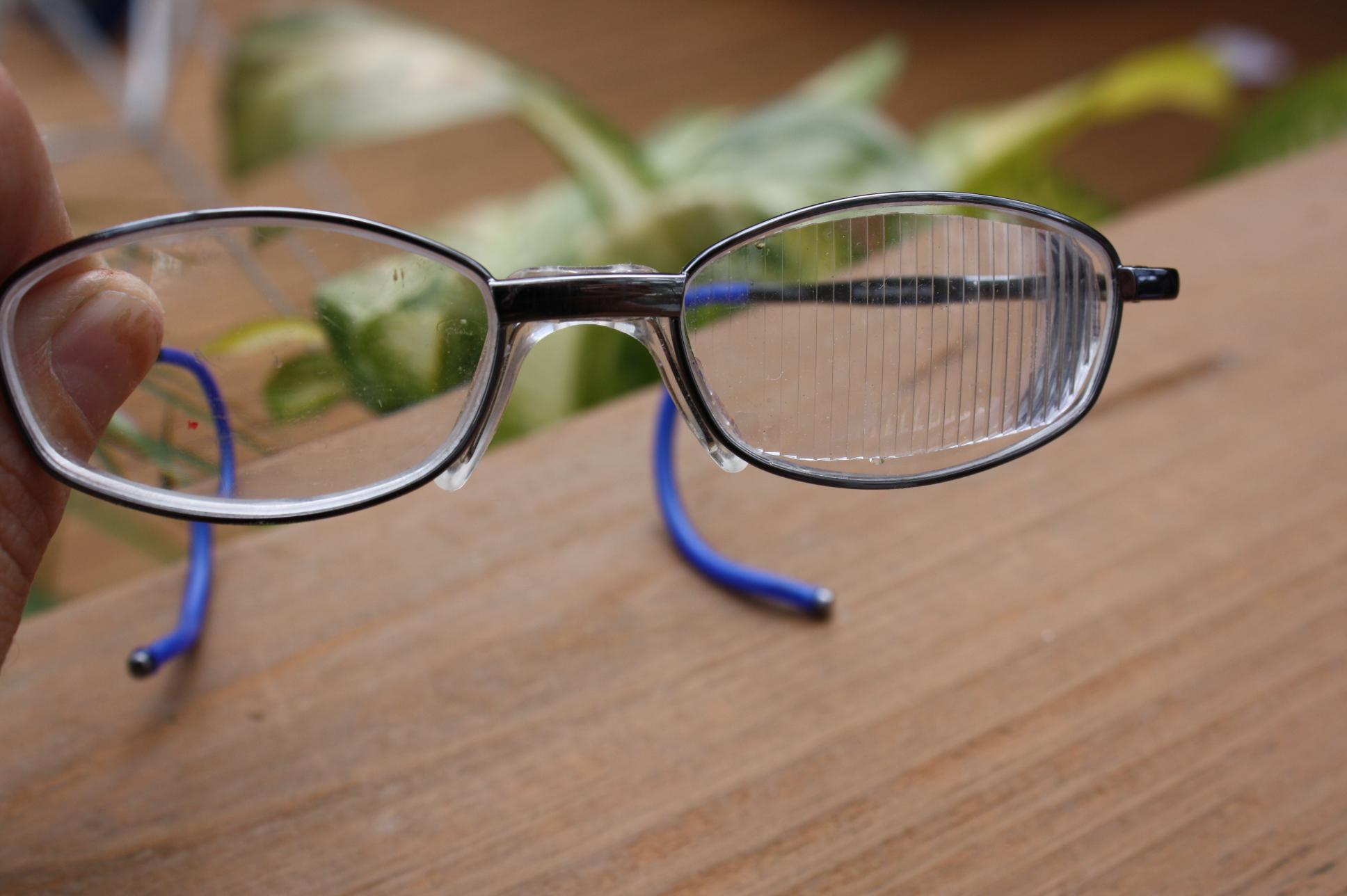}    
\caption{Glasses}
\end{subfigure}
\begin{subfigure}[b]{.15\columnwidth}
\includegraphics[width=1.4cm,height=1.4cm,keepaspectratio]{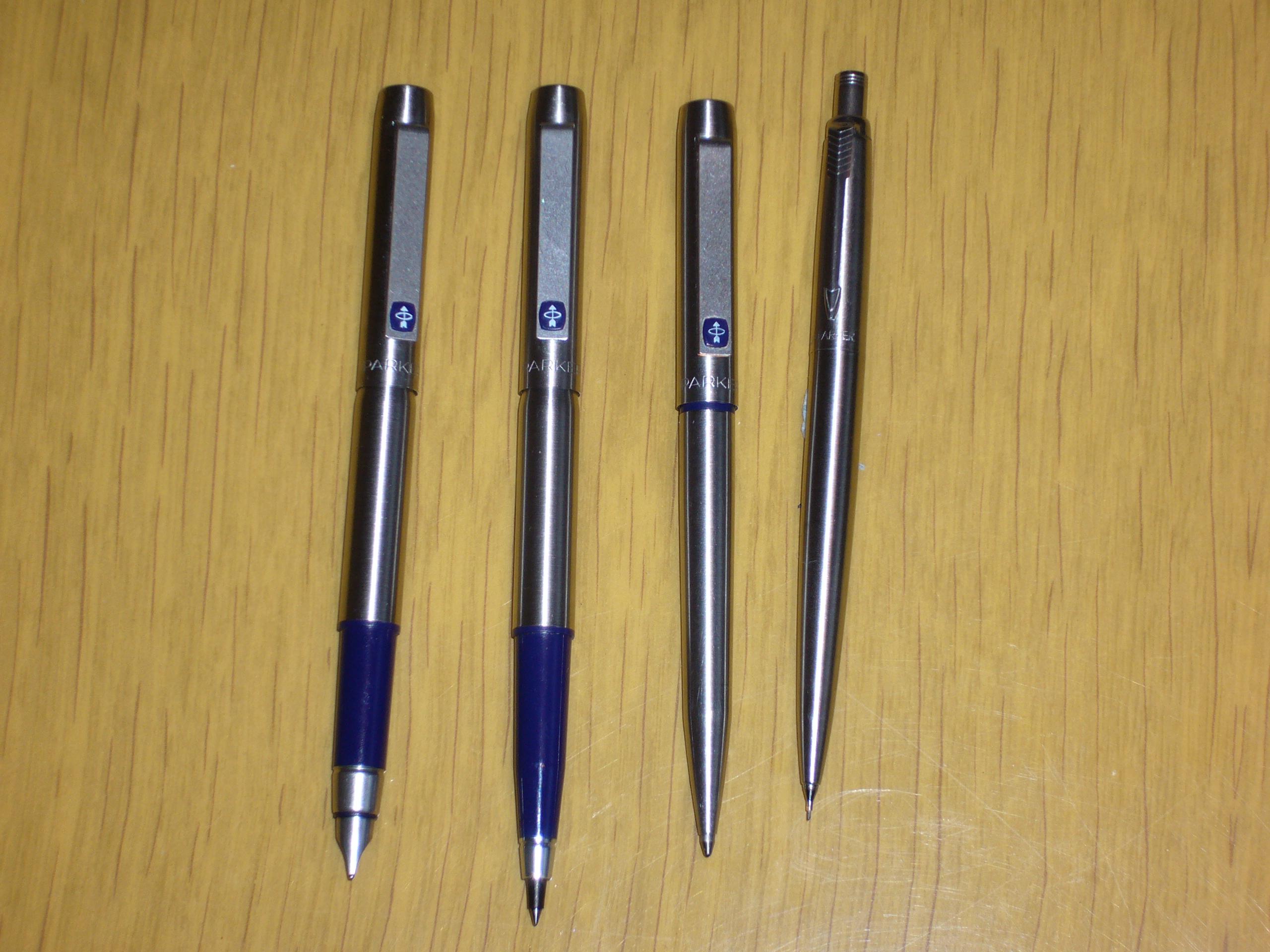}      
\caption{Pen}
\end{subfigure}
\begin{subfigure}[b]{.15\columnwidth}
\includegraphics[width=1.4cm,height=1.4cm,keepaspectratio]{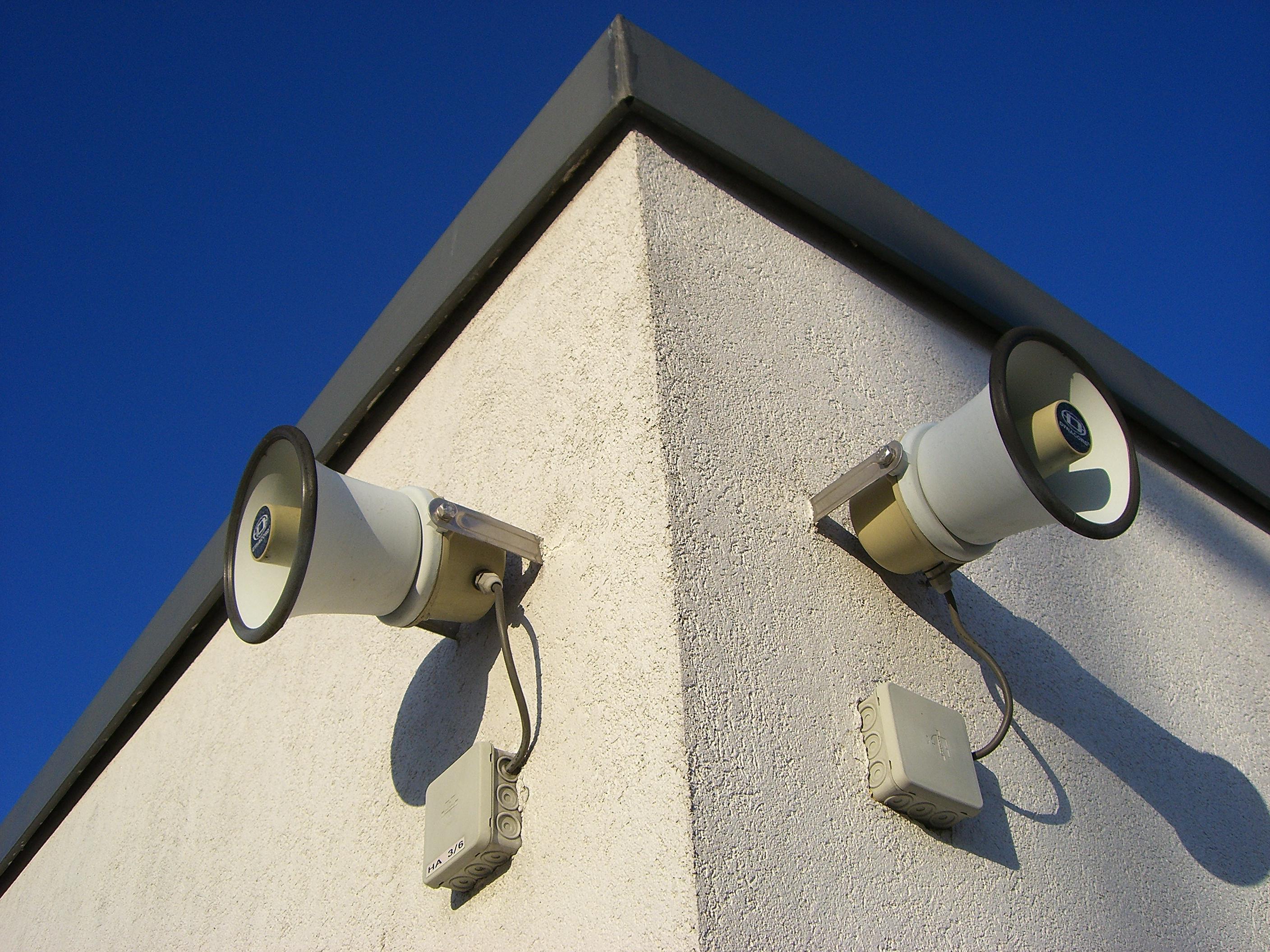}
\caption{Speaker}
\end{subfigure}
\caption{Example images from $5$ of the $65$ categories from the $4$ domains composing the Office-Home dataset \cite{venkateswara2017Deep}. The $4$ domains are Art, Clipart, Product and Real World. In total, the dataset contains $15{,}500$ images of different sizes.}
\label{fig:Dataset}
\end{figure}

\begin{align}\label{prob:MTL_agnostic_EMP}
   \hat P_{A}^*=\underset{\{\theta_t\}}{\min}  \quad   &\sum_{t=1}^T \frac{1}{N_t} \sum_{i=1}^{N_t} \ell\left(y_i,f(x_i,\theta_t)\right).
\end{align}
Note that the multi-task learning problem \eqref{prob:MTL_agnostic_EMP} is separable over the tasks, which implies that we can obtain each solution $\theta_t$ by solving $T$ optimization problems independently. In the multi-task learning setup however, data from different tasks may be correlated. Therefore, solving \eqref{prob:MTL_agnostic_EMP} with a complete separation between tasks could possibly render a loss of relevant information. Before proceeding, we will exemplify equation \eqref{prob:MTL_agnostic_EMP} with Figure \ref{fig:Dataset}. For example, in an image classification problem, the goal can be to correctly identify images of alarms. In the case of multi-task learning, images might also come from different sources, this is the case in Figure \ref{fig:Dataset}, where there are images of artistic representations of alarms (Art), as well as images of alarms as shown for sale (Product), or images taken in with a camera (World). Intuitively, by looking at the images, we can conclude that images of the same class are related between domains. 

An immediate approach to leverage information between tasks is to impose a common solution $\theta$ for all tasks as follows, 
\begin{align}\label{prob:MTL_centralized}
   \hat P_{C}^*= \underset{\theta}{\min}  \quad   &  \frac{1}{\sum_{t=1}^TN_t}\sum_{t=1}^T \sum_{i=1}^{N_t} \ell\left(y_i,f(x_i,\theta)\right).
\end{align}
By combining the data from all tasks, problem \eqref{prob:MTL_centralized} seeks for a unique common solution. Note that the summation is taken over all samples, i.e., $\sum_{t=1}^TN_t$, and therefore the dependency on the tasks is removed. Albeit related, tasks tend to differ from each other, and enforcing a single function for all tasks might be too restrictive, possibly resulting in a loss of optimality. 
\subsection{Multi-Task Learning Bias-Variance Trade-off}
In this section we will argue that there is a bias-variance trade-off between estimating the solutions of the multi-task learning problem \eqref{prob:MTL_agnostic} via the agnostic estimator  \eqref{prob:MTL_agnostic_EMP}, versus the centralized estimator \eqref{prob:MTL_centralized}. For this purpose, we need to introduce the following probably approximately correct assumption, as well as an assumption on the smoothness of the loss function $\ell$,
\begin{assumption}\label{ass:pac}
Function $f(\cdot,\theta)$ is probably approximately correct (PAC), i.e. there exists $\zeta(N,\delta)\geq 0$, monotonically decreasing with $N$, such that, for all $\theta\in\Theta$, with probability $1-\delta$ over independent draws $(x_n,y_n)\sim p$,
\begin{align}
    \bigg|\mbE[\ell(f(x,\theta),y)]-\frac{1}{N}\sum_{i\in[N]} \ell(f(x_n,\theta),y_n) )\bigg|\leq \zeta(N,\delta)
\end{align}
\end{assumption}
\begin{assumption}\label{assumption:convex_function}
Function $\ell(\cdot,y)$ is $M$-Lipschitz continuous for all $y$.  
\end{assumption}

 Assumption \ref{ass:pac}   comes as a generalization of the law of large numbers for the case in which samples are i.i.d., where the error is of order $1/N$. Moreover, this generalized version of the empirical error can be found in terms of the Rademacher complexity, as well as the Vapnik-Chervonenkis dimension \cite{vapnik1999nature,mohri2018foundations}.
Under this PAC assumption we are able to provide the following proposition,
\begin{proposition}\label{prop:learning_tradeoff}
Under Assumptions \ref{ass:pac}-\ref{assumption:convex_function} with probability at least $1-T\delta$ the following bounds hold,
\begin{align}
    |P_A^* - \hat P_{A}^* | \leq& \sum_{t=1}^T\zeta\big( N_t,\delta\big)\label{eqn:bound_learning_tradeoff_ags}\\
   | P_A^* - \hat P_{C}^* | \leq&  M\sum_{t=1}^T \mbE_{p_t}[|f(x,\theta^*_t)-f(x,\theta^*_g)|]  \nonumber\\&+  \zeta\big(\sum_{t=1}^T N_t,\delta\big)\label{eqn:bound_learning_tradeoff_cen}
\end{align}
where $\{\theta^*_t\}$ are the solutions to problem \eqref{prob:CL}, and $\theta^*_g$ is the solution to the statistical version of \eqref{prob:MTL_centralized}. 
\end{proposition}
\begin{proof}
By a direct application of the triangle inequality, and Assumption \ref{ass:pac} we get \eqref{eqn:bound_learning_tradeoff_ags}.
In the case of \eqref{eqn:bound_learning_tradeoff_cen}, note that it is also a direct application of Assumption \ref{ass:pac} over the average probability distribution, i.e. $\bar p(x,y) = \sum_{t=1}^T p_t(x,y)$, and taking the Lipschitz Assumption \ref{assumption:convex_function} over the solutions.
\end{proof}
Proposition \ref{prop:learning_tradeoff} bounds the difference between the solution of the multi-task learning problem  \eqref{prob:MTL_agnostic}, and its empirical estimator \eqref{prob:MTL_agnostic_EMP}, and the centralized empirical estimator \eqref{prob:MTL_centralized}. In the case of the agnostic bound \eqref{eqn:bound_learning_tradeoff_ags},  we can conclude that the empirical estimator \eqref{prob:MTL_agnostic_EMP} is unbiased, and that variance of the estimator is given by the summation of $\zeta(N_t,\delta)$. Whereas in the case of the centralized empirical estimator \eqref{eqn:bound_learning_tradeoff_cen}, the bound has two terms, one term that depends on the number of samples, and another one that depends on the distance between the solutions, $\mbE_{p_t(x,y)}[|f(x,\theta^*_t)-f(x,\theta_g^*) |]$. The distance between the solutions can be thought of a bias term, i.e., a term that does not reduce if the number of samples increase. Given that $\zeta$ tends to be sub-linear, $\sum_t\zeta(N_t,\delta)\geq\zeta(\sum_t N_t,\delta)$, and therefore the variance term of the centralized estimator will be smaller than its agnostic counterpart. 
The distance between the solutions in \eqref{eqn:bound_learning_tradeoff_cen} is unknown and cannot be computed in practice, however it serves as a guideline to design estimators that allows information between tasks to be shared, and control the bias term they introduce by doing so.

\subsection{Cross-Learning}
In order to exploit the bias-variance trade-off presented in Proposition \ref{prop:learning_tradeoff}, we introduce a novel cross-learning formulation.
\begin{align}
    P_{CL}^*=\underset{\{\theta_t\},\theta_g}{\min}  \quad   &  \sum_{t=1}^T\mbE_{p_t(x,y)}[ \ell \left(y,f(x,\theta_t)\right)] \label{prob:CL}
 \\
   \text{subject to}
      \quad   & \mbE_{p_t}[| f(x_t,\theta_t)-f(x_t,\theta_g)|]\leq \epsilon.\nonumber
\end{align}
The cross-learning problem \eqref{prob:CL} connects both the multi-task learning problem \eqref{prob:MTL_agnostic}, and the centralized problem \eqref{prob:MTL_centralized} by enforcing a constraint on the closeness between functions. By forcing functions to be close to the central one, we relax the centralized problem \eqref{prob:MTL_centralized} allowing for solutions of different tasks to differ from each other. The centrality constant $\epsilon$ controls the closeness between functions, selecting a larger $\epsilon$ allows functions to be more task-specific.
Note that by enforcing $\epsilon=0$, all functions are required to be equal over $p_t$, and problem \eqref{prob:CL} becomes equivalent to \eqref{prob:MTL_centralized}. Likewise, if $\epsilon$ is sufficiently large, the constraint in \eqref{prob:CL} is rendered inactive, and problem \eqref{prob:CL} becomes equivalent to \eqref{prob:MTL_agnostic}.

The advantages of solving problem \eqref{prob:CL} with respect to both the centralized, and task-specific problems is two-fold (i) as opposed to the agnostic problem \eqref{prob:MTL_agnostic_EMP}, we combine the information from one task to another through the constraint and therefore we exploit data from different tasks, (ii) as opposed to \eqref{prob:MTL_centralized} we allow functions associated with different tasks to be different to each other, therefore allowing a better task-specific performance.  Returning to Proposition \ref{prop:learning_tradeoff}, cross-learning intends to control the bias-variance trade-off by enforcing the task-specific solutions to be close. 

\subsection{Functional Constraints}
Enforcing constraints over expected values of functions $f(\cdot,\theta)$ cannot be computed given that we do no have access to the probability distribution $p_t$. A surrogate of the distance between functions can be to utilize the distance between parameters of the function, i.e. $|\theta_t-\theta_g|$ \cite{cervino2020multi}. 
Note that if the function is Lipschitz in the parameterization, there is a connection between the functional, and parametric constraints, 
\begin{proposition}\label{prop:FunctionalParametric}
If the parametric function is Lipschitz in the parameterization i.e. $|f_{\theta_1}(x)-f_{\theta_2}(x)|\leq L |\theta_1-\theta_2| $ then, 
\begin{align}
    \mbE[| f(x_t,\theta_t)-f(x_t,\theta_g)|]\leq L |\theta_t-\theta_g|.
\end{align}
\end{proposition}

Note that enforcing the constraint over the parameters makes the constraint convex, and removes the expectation, and therefore dependency over the distribution $p_t$. 
However, in the case of neural networks, constraining the parameters might be too restrictive, as different sets of parameters might evaluate to the same function.
Moreover, given the layered architecture of neural networks, a small difference in the first layers might be propagated throughout the network, making even small differences in parameters have large deviations in the resulting outputs.
The functional constraint is more interpretable than its parameter counterpart, as in the case of classifiers where the output is a probability vector, and the norm of the difference between functions can be interpreted as a difference between probabilities distributions over the classes.

\section{Dual Domain}
\label{sec:DualDomain}
Upon presenting the cross-learning problem \eqref{prob:CL} as a formulation to better balance the variance-bias trade-off, in this section we propose a method to solve it. 
Note that solving problem \eqref{prob:CL} cannot be done in practice given that the distributions $p_t$ are unknown. 
Therefore, we leverage recent results in the area of stochastic dual optimization to develop a primal-dual methodology to solve the problem \cite{chamon2020empirical,chamon2020probably}. Let $\lambda_t>0$ be the dual variable associated with domain $t$, and $ \lambda=[\lambda_1,\dots,\lambda_T]^T$,  we define the Lagrangian associated with the cross-learning problem \eqref{prob:CL} as
\begin{align}
 \hat L(\theta_t,\theta_g,\lambda):=& \sum_{t=1}^T \frac{1}{N_t} \sum_{i=1}^{N_t} \ell \left(y_i,f(x_i,\theta_t)\right) \nonumber\\ &+\lambda_t\bigg(\frac{1}{N_t} \sum_{i=1}^{N_t}| f(x_t,\theta_t)-f(x_t,\theta_g)|- \epsilon\bigg).
\end{align}
We define the dual function associated as follows,
\begin{align}
    d(\lambda) := \min_{\theta_t,\theta_g}\hat L (\theta_t,\theta_g,\lambda). \label{eqn:dual_function}
\end{align}
The dual function $d(\lambda)$ is a concave function over $\lambda$ \cite[Section 5.1.2]{boyd2009convex}. What is more, for a given value of $\lambda$, finding the dual function entails solving an unconstrained problem over $\theta$. We can now define the dual problem associated with the cross-learning problem \eqref{prob:CL} as follows,
\begin{algorithm}[t]
	\caption{Cross-Learning Functional Algorithm}
	\label{alg:Algorithm}
	\begin{algorithmic}[1]
	\State Initialize models $\{\theta^0_t\},\theta^0_g$, and dual variables $\lambda = 0$
    \For {epochs $e=1,2,\dots$}
    \For {batch $i$ in epoch $e$}
    \State Update params. $\theta_t^{k+1}=\theta_t^{k}-\eta_P \hat\nabla_{\theta_t} L (\theta_t,\theta_g,\lambda)\forall \ t$
    \State Update $g$ params. $\theta_g^{k+1}=\theta_g^{k}-\eta_P \hat\nabla_{\theta_g} L (\theta_t,\theta_g,\lambda)$
    \EndFor
    \State Update dual variable for all $t\in[1,\dots,T]$ \quad~\quad~\quad~\quad~\quad~$\quad$ $\lambda_t^{k+1} = \biggl[\lambda^{k}_t + \eta_D\bigg(\frac{1}{N_t} \sum_{i=1}^{N_t}| f(x_t,\theta_t)-f(x_t,\theta_g)|- \epsilon \bigg)\biggl]_{+}$ 
    \EndFor
	\end{algorithmic}
\end{algorithm}
\begin{align}\label{prob:dual_CL}
    D^*_{CL} :=\max_{\{\lambda_t\}} d(\lambda) 
\end{align}
The dual problem $D_{CL}^*$ is obtained by taking the maximum over $\lambda$ of the dual function $d(\lambda)$, and it is a convex optimization problem. 

\subsection{Algorithm Construction}
\label{subsec:algo}

In order to solve the cross-learning dual problem \eqref{prob:dual_CL}, we will implement an iterative primal-dual algorithm. Upon initializing the parameters $\{\theta_t^0\},\theta_g^0$, and the dual variables $\lambda_t^0$, we take a gradient step to minimize the dual function \eqref{eqn:dual_function} as follows,
\begin{align}
\theta^{k+1}_t = \theta^k_t -\eta_P \nabla_\theta \hat L(\theta_t,\theta_g,\lambda)\text{ for all } t\in[1,\dots,T] \text{ and }g,
\end{align}
where $\eta_P>0$ is the step size.  Upon taking steps to minimize the dual function \eqref{eqn:dual_function}, we update the value of the dual variable by evaluating the constraint satisfaction as follows,
\begin{align*}
\lambda^{k+1}_t = \bigg[\lambda^{k}_t + \eta_D\bigg(\frac{1}{N_t} \sum_{i=1}^{N_t}| f(x_t,\theta_t)-f(x_t,\theta_g)|- \epsilon \bigg)\bigg]_+
\end{align*}
where $\eta_D>0$ is the dual step-size, and $[\cdot]_+=\max\{0,\cdot\}$. The overall primal-dual procedure is summarized in Algorithm \ref{alg:Algorithm}. It can be shown that Algorithm \ref{alg:Algorithm} converges \cite{chamon2020probably}, but this proof is out of the scope of this paper.

\section{Experiments}
\label{sec:Experiments}

In this section we benchmark our cross-learning framework on a classification problem with real data coming from the dataset that we introduced in Figure \ref{fig:Dataset}. We consider the problem of classifying images belonging to $P=65$ different categories (which in Figure \ref{fig:Dataset} are Alarm, Bike, Glasses, Pen, and Speaker), and $N=4$  different domains. The Office-Home dataset \cite{venkateswara2017Deep} consists of $15{,}500$ RGB images in total coming from the domains (i.e. tasks) (i) Art: an artistic representation of the object, (ii) Clipart: a clip art reproduction, (iii) Product: an image of a product for sale, and (iv) Real World: pictures of the object captured with a camera. Intuitively, by looking at the images, we can conclude that the domains are related. The minimum number of images per domain and category is $15$ and the image size varies from the smallest image size of $18 \times 18$ to the largest being $6500 \times 4900$ pixels. We pre-processed the images by normalizing them and fitting their size to $224 \times 224$ pixels.

\begin{figure}[t]
\centering
\includegraphics[width=0.975\columnwidth]{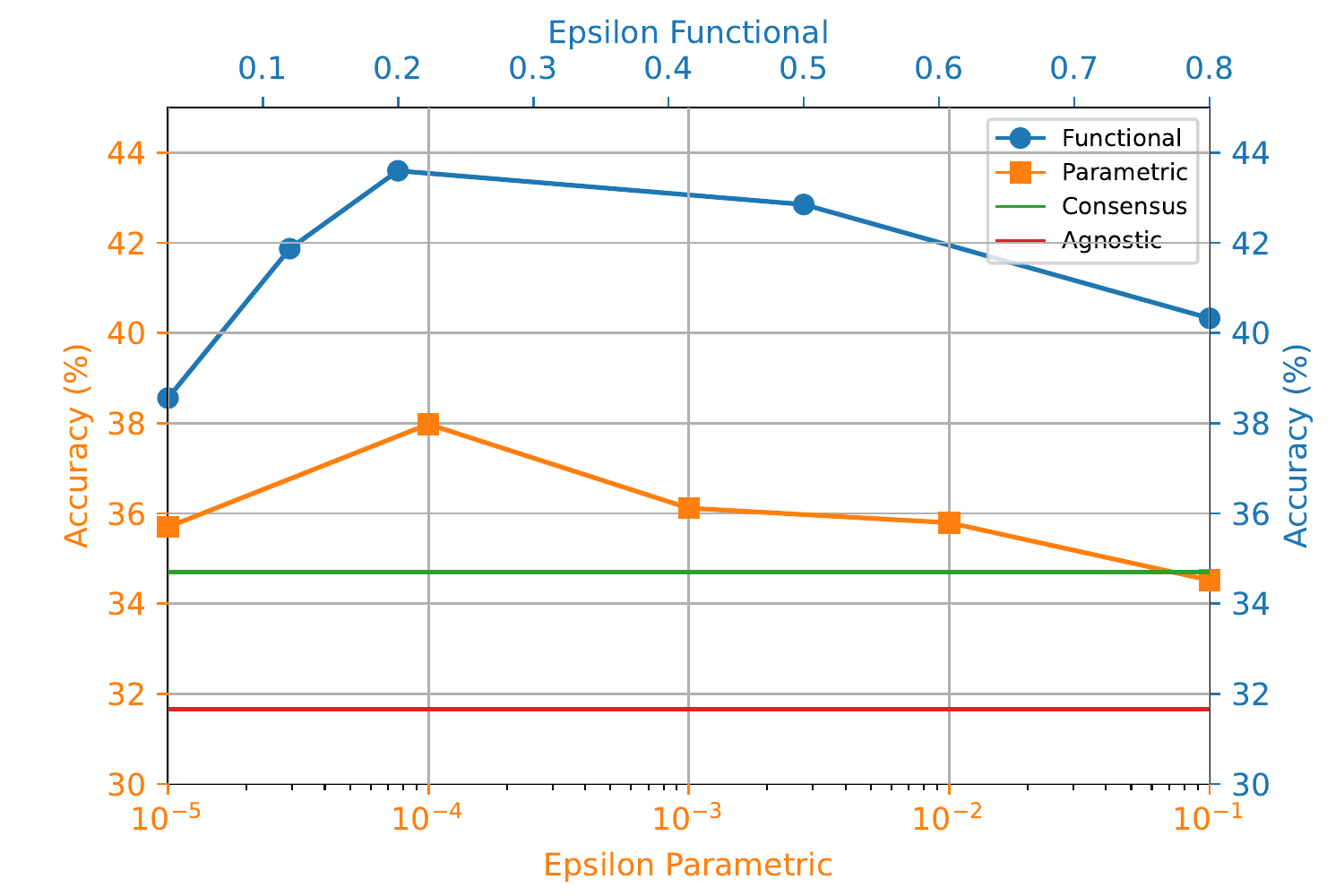}
\caption{Overall accuracy in percentage of correctly classified images measured on unseen data (test set). The consensus case corresponds to $\epsilon=0$ and the agnostic case corresponds to $\epsilon=\infty$. }
\label{fig:accuracy}
\end{figure}

For the classifiers, we use neural networks $f(x,\theta_i)$, $i=1,\ldots,N$ with the architecture being based on AlexNet \cite{krizhevsky2012imagenet} with a reduction on the size of the last fully connected layer to $256$ neurons, and we do not use pre-trained networks. For the loss, we use the cross-entropy loss \cite{hastie2009elements}, and set as primal, and dual step-size  $\eta_P=0.003$, and $\eta_D=10$ respectively. Furthermore, we split the dataset in two parts, using  $4/5$ of the images  for training and $1/5$ for testing. 

For our Algorithm \ref{alg:Algorithm}, we train $5$ neural networks with different values of the centrality measure $\epsilon$. We compare our algorithm against the consensus classifier ($\epsilon=0$) -- which is equivalent to merging the images from all domains and training a single neural network on the whole dataset-- and against the $4$ agnostic classifiers ($\epsilon=\infty$) which is equivalent to training each neural networks separately in each domain. We utilize as a benchmark the cross-learning algorithm with parameter constraint (cf. Proposition \ref{prop:FunctionalParametric}) \cite{cervino2020multi}.

We show our results in Figure \ref{fig:accuracy}, where we use the classification accuracy of the trained classifiers on the unseen data (test set) as figure of merit.  
To begin with, we empirically corroborate that the domains are correlated as the consensus classifier ($\epsilon=0$) outperforms the agnostic training ($\epsilon=\infty$) (accuracy of $33.97\%$ against $24.44\%$). This means that utilizing samples from other domains improves the accuracy of the learned classifier. This is the setting in which cross-learning is most helpful as it will help the reduce the bias term (cf. Proposition \ref{prop:learning_tradeoff}).

The advantage of cross-learning can be better seen in Figure \ref{fig:accuracy}, as for both parametric \cite{cervinoeusipco}, and functional (this work) cross-learning there exists a value of $\epsilon$ that outperform both the consensus, and agnostic classifiers. Moreover, the advantage of functional constraints can be seen as the functional version (blue) obtains a better accuracy than the parametric version (orange). In all, the functional version of cross-learning outperforms both the agnostic, and consensus classifiers, as well as the parametric baseline. In Figure \ref{fig:accuracy} the functional (blue), and parametric (orange) have different scales for the value of the proximity $\epsilon$, this is due to the the fact that even related the constraints are enforced on different spaces. 

\begin{figure}[t]
\centering
\includegraphics[width=0.975\columnwidth]{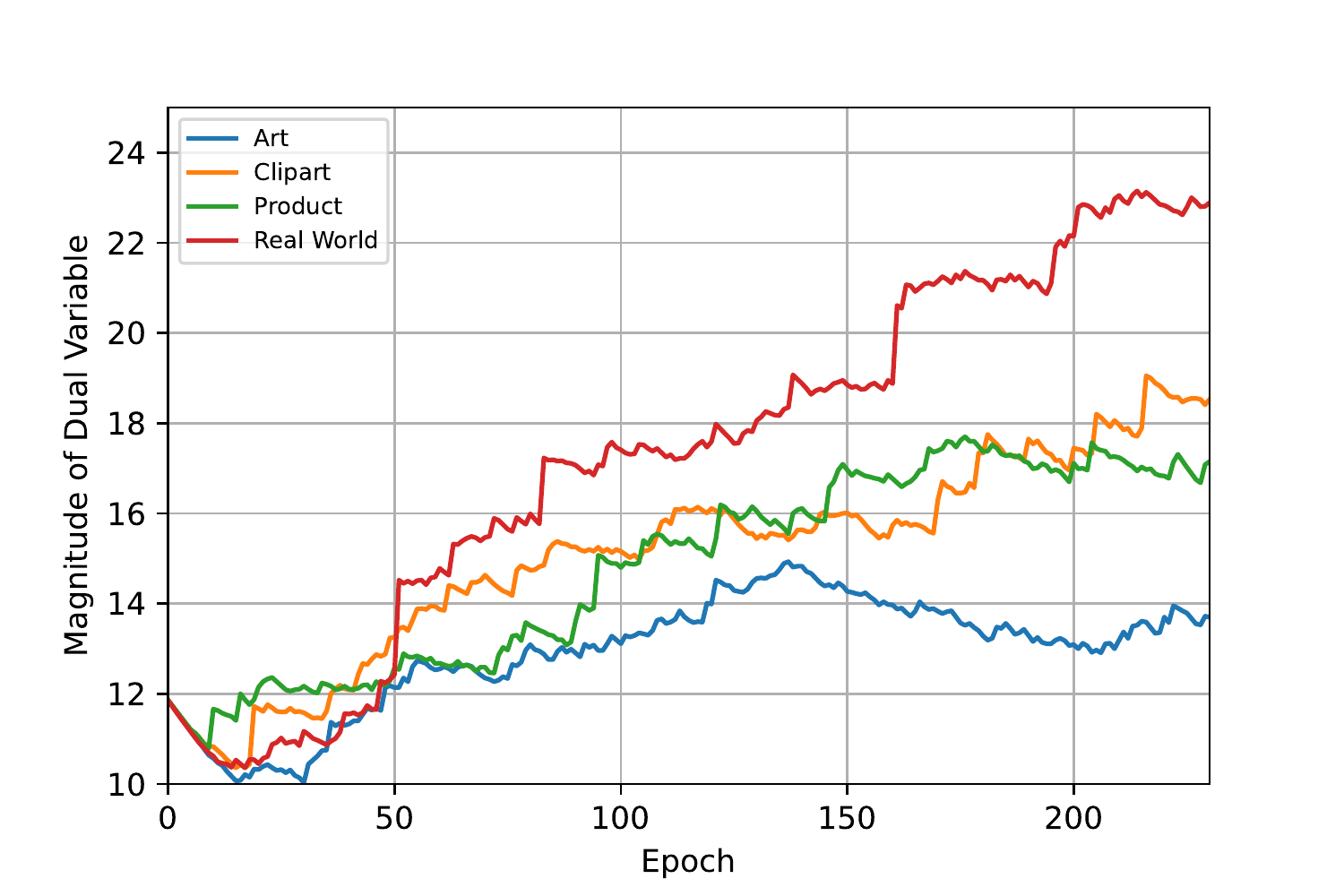}
\caption{Dual variable associated with each constraint for the case of $\epsilon=0.12$. Intuitively, a larger dual variable indicates that the constraint is harder to satisfy.}
\label{fig:lambdas}
\end{figure}

As a byproduct, Algorithm \ref{alg:Algorithm} generates dual variables $\lambda_t$ that contain valuable information of the problem at hand. In Figure \ref{fig:lambdas} we plot the value of the dual variables $\lambda$ as a function of the epoch. A larger dual variable indicates that the constraint is harder to satisfy, and a dual-variable equal to $0$ indicates that the constraint is inactive. As seen in \ref{fig:lambdas}, all $\lambda_t$ are non-negative, which means that we are effectively in a regime where there the classifiers are utilizing data from other tasks. 
If we look at the relative values of the dual variables, we see that the domain Art has the smallest value, whereas the domain Real World has the largest one. Given that Art has the least amount of samples, it is the domain that mostly benefits from images coming from other tasks. On the other hand, the largest dual variable is associated with the real-world dataset, which is can be explain by the fact that the images it posses have more details, textures, and shapes, and are therefore more difficult to classify (cf. Figure \ref{fig:Dataset}).


\section{Conclusion}
In this paper we presented a constrained learning approach to jointly learn functions belonging to different domains. 
We proposed a multi-task learning problem with constraints on the difference between the task specific classifiers, and a central classifier for all domains. To solve this optimization problem utilizing samples coming from the task distributions,  we proposed a primal-dual algorithm that iteratively updates the value of the functions, as well as the dual variables associated with each constraint. 
We bench-marked our procedure in a classification problem with real data coming from different domains.
\newpage

\bibliographystyle{IEEEbib}
\bibliography{bib}
\appendix

\end{document}